\documentclass[runningheads]{llncs}
\usepackage{graphicx}
\usepackage{amsmath,amssymb} 
\usepackage{color}
\usepackage[width=122mm,left=12mm,paperwidth=146mm,height=193mm,top=12mm,paperheight=217mm]{geometry}

\usepackage{url}
\usepackage{diagbox}
\usepackage{multirow}
\usepackage{multicol}
\usepackage{tablefootnote}

\input{latex.macro}
\graphicspath{{images/}}

\begin{document}

\mainmatter

\title{Shared Representation Learning for Heterogeneous Face Recognition} 



\author{Dong Yi, Zhen Lei, Shengcai Liao and Stan Z. Li}
\institute{National Laboratory of Pattern Recognition,
Institute of Automation, Chinese Academy of Sciences, Beijing, China, 100190}

\maketitle

\begin{abstract}
After intensive research, heterogenous face recognition is still a challenging problem.
The main difficulties are owing to the complex relationship between heterogenous face
image spaces. The heterogeneity is always tightly coupled with other variations,
which makes the relationship of heterogenous face images highly nonlinear. Many excellent
methods have been proposed to model the nonlinear relationship, but they
apt to overfit to the training set, due to limited samples.
Inspired by the unsupervised algorithms in deep learning, this paper
proposes an novel framework for heterogeneous face recognition. We first
extract Gabor features at some localized facial points, and then use Restricted Boltzmann Machines (RBMs)
to learn a shared representation locally to remove the heterogeneity around each facial point.
Finally, the shared representations of local RBMs are connected together and processed by
PCA. Two problems (Sketch-Photo and NIR-VIS) and three databases are selected to evaluate
the proposed method. For Sketch-Photo problem, we obtain perfect results on the CUFS database.
For NIR-VIS problem, we produce new
state-of-the-art performance on the CASIA HFB and NIR-VIS 2.0 databases.

\end{abstract}

\section{Introduction}

The core of heterogenous face recognition~\cite{Li-Biometrics-2009} is face matching across modalities. Although the original definition of heterogenous face recognition is broad, the two hottest problems about this topic are Sketch-Photo~\cite{Wang-TPAMI-2009} face recognition and NIR-VIS (Near Infrared-Visual)~\cite{Yi-NIR-VIS-ICB-07} face recognition. This paper will also take these two problems as examples to verify the proposed method.

Initially, heterogeneous face recognition was proposed to appeal requirements in practical applications. Sketch-Photo matching is often required in law enforcement when the photo of suspect
is unavailable. NIR-VIS matching module can make VIS face recognition system work
in dark environment using NIR imaging device. After several research groups were attracted to this
topic, many good methods have been proposed and these methods quickly spread to other cross-modal
problems, such as face hallucination~\cite{Wang-TSMC-2005}, pedestrian detection~\cite{Yan-CVPR-2013} and so on.

It has been shown in existing works that the relationship of face images between different modalities
is very complex, therefore nonlinear methods usually have better performance than linear methods.
Taking NIR-VIS as an example, the effect of spectrum is tightly coupled with other variations of
face image, such as 3D shape, pose, identity and so on, which makes the relationship of face images
under different spectrums highly nonlinear and varying with respect to locations. Among existing
methods, the most successful category is learning two mappings (linear or nonlinear) to project
the heterogenous face images into a common space~\cite{Lin-ECCV-06}\cite{Lei-TIFS-2012}. Limited by the number of training samples, this kind of methods have many regularization terms, so need careful parameter tuning to achieve good
performance.

From 2006 to now, unsupervised pre-training has obtained great success in deep learning~\cite{Hinton-SCI-2006}. One of the
most popular unsupervised learning method in deep learning is Restricted Boltzmnn Machine (RBM)~\cite{Smolensky-1986},
which is a generative stochastic neural network that can learn a probability distribution of input data.
To improve the generalization of existing methods and make the training process easily, this paper propose a framework based on RBM to learn the relationship of face images between different modalities. Because RBM is nonlinear and unsupervised,
our framework can learn the nonlinear relationship well and unlikely prone to overfitting.

The proposed framework includes 3 main steps: (1) extracting local Gabor features around facial points, as
traditional face recognition methods do; (2) learning a shared representation by RBM for each group of
local features; (3) processing the whole RBM representations by PCA and matching by Cosine similarity. Among them the key
step is (2), in which a 3-layer RBM is constructed and the middle layer represents the shared properties
of heterogeneous data.

The contributions of this paper are as follows.
\begin{enumerate}
\item A local to global learning framework is proposed for heterogeneous face recognition, which can achieve good results in all experiments.
\item For Sketch-Photo problem, perfect results are obtained on the CUFS~\cite{Wang-TPAMI-2009} database. This is the first work that saturates the database.
\item Local RBMs are first used to learn the shared representations of heterogenous face images. By plugging the local RBMs into the framework, we get state-of-the-art results on the CASIA HFB~\cite{Li-CVPRW-2009} and NIR-VIS 2.0~\cite{Li-CVPRW-2013} databases.
\end{enumerate}

\section{Related Works}
\label{sec:review}

Heterogeneous face recognition research started from Tang and Wang's work in 2002~\cite{Tang-ICIP-2012}. From that time to now, existing methods can be divided into two categories: Synthesis based and Classification oriented
methods. In the early stage, the mainstream belongs to synthesis based methods, such as \cite{Tang-CVPR-2003}, \cite{Liu-CVPR-05} and \cite{Wang-ICB-2009}. \cite{Tang-CVPR-2003} proposed a method, named as eigen-transformation, to synthesize photo by sketch and then recognized the identity in photo modality. To get more realistic results, \cite{Liu-CVPR-05} synthesized photo in a patch way, in which each image patch was first reconstructed by LLE and then stitched into a whole photo. \cite{Wang-ICB-2009} also proposed a simple way to transform VIS to NIR face image. Although the results show that synthesis based method can achieve good visual quality, the recognition rate based on the synthesized images is moderate.

In late years, more classification oriented methods were proposed to improve the recognition rate directly.
These methods just have one target: removing the difference of modalities, and meanwhile
extracting discriminative feature. Many image processing and coding techniques are their essential parts,
such as DoG filter~\cite{Liao-ICB-2009}, LBP, HOG~\cite{Klare-TPAMI-2011}, using which the difference between Sketch, NIR or VIS face images can be reduced significantly. Then, the processed heterogenous data are mapped to a discriminative space by linear, nonlinear mapping~\cite{Lin-ECCV-06}\cite{Lei-TIFS-2012} or random trees~\cite{Zhang-CVPR-2011}. Because the target of this kind of methods is more direct than synthesis based methods, they always perform better.

Recently, several methods are proposed for multi-modal problems in deep learning community. \cite{Ngiam-ICML-2011} first proposed a multi-modal deep learning method based on denoising autoencoder, named as Bimodal Deep AE. But the Bimodal Deep AE performs poorly in Video-Audio matching experiments. On the contrary, another shallow architecture RBM-CCA results in surprisingly good performance. Unfortunately, \cite{Ngiam-ICML-2011} didn't give any analysis about why the deep net was worse than RBM-CCA. In 2012, \cite{Srivastava-NIPS-2012} pointed out that in Bimodal Deep AE the responsibility of the multi-modal modeling fell entirely on the joint layer, and other layers gave no contributions. Therefore, they proposed a multi-modal Deep Boltzmann machine (DBM), which can spread out the responsibility of the multi-modal modeling over the entire network. Experiments illustrated the superiority of DBM in Image-Text retrieval task. Then, \cite{Feng-2013} applied the multi-modal DBM in the Image-Text retrieval challenge of ICML 2013 and got the first place in the challenge.

Because the multi-modal RBM in \cite{Srivastava-NIPS-2012} has many good properties to deal with cross-modal matching problem, we plug the multi-modal RBM into the traditional face recognition pipeline to construct a novel framework for
heterogeneous face recognition. By combing the advanced modules in these two fields, the proposed framework can work very well in challenging experiments.

\section{Background}
\label{sec:bg}

RBM has been widely used for modeling distribution of binary data. After Hinton's work~\cite{Hinton-SCI-2006}, it became a
standard building block of deep neural network. To model the real-valued data of face images, Gaussian RBM is used in this paper. This section will review the RBM, Gaussian RBM and Multi-modal RBM in brief.

\subsection{Restricted Boltzmann Machines}

RBM~\cite{Smolensky-1986} is a generative stochastic graphical model that can learn the distribution of training data. The model consists of stochastic visible units $\bfv \in \{0, 1\}^m$ and stochastic hidden units $\bfh \in \{0, 1\}^n$,
which aims to minimize the following energy function:
\begin{equation}
E(\bfv, \bfh; \bfa, \bfb, \bfW) = -\bfa^T \bfv - \bfb^T \bfh - \bfv^T \bfW \bfh,
\end{equation}
where $\bfa$ is the biases of visible units; $\bfb$ is the biases of hidden units; $\bfW$ is the weights matrix
to connect the visible and hidden units.

For image data, real-valued visible units $\bfv \in \mathbb{R}^m$ are used to replace the binary ones. The new model is called Gaussian RBM~\cite{Hinton-Guide-2012}, the energy function of which is defined as:
\begin{equation}
E(\bfv, \bfh; \bfa, \bfb, \bfW) = \frac{1}{2} \bfu^T \bfu - \bfb^T \bfh - (\bfv \odot \frac{1}{\boldsymbol\sigma})^T \bfW \bfh,
\end{equation}
where $\bfu=(\bfv - \bfa) \odot \frac{1}{\boldsymbol\sigma}$ denotes the normalized visible data. $\boldsymbol\sigma$ is a vector consisting of the standard deviations of each dimension. $\odot$ denotes element-wise multiplication of vectors. Before training Gaussian RBM, the input data are usually normalized by WPCA or ZCA~\cite{Bell-VR-1997}, \ie the standard deviations $\boldsymbol\sigma$ of the normalized data $\hat{\bfv}$ is 1. Then, the energy function can be simplified as:
\begin{equation}
E(\hat{\bfv}, \bfh; \bfa, \bfb, \bfW) = \frac{1}{2} (\hat{\bfv} - \bfa)^T (\hat{\bfv} - \bfa) - \bfb^T \bfh - \hat{\bfv}^T \bfW \bfh.
\end{equation}

Then the distribution over visible and hidden units is defined as:
\begin{equation}
P(\hat{\bfv}, \bfh; \boldsymbol\theta) = \frac{1}{Z} e^{-E(\hat{\bfv}, \bfh; \boldsymbol\theta)},
\label{equ:energy-to-prob}
\end{equation}
where $\boldsymbol\theta$ is an abberation for the parameters of RBM $\{\bfa, \bfb, \bfW$\}; $Z$ is a partition function defined as the sum of $e^{-E(\hat{\bfv}, \bfh; \boldsymbol\theta)}$ over all possible configurations.

\subsection{Multi-modal RBM}

\begin{figure}
  \centering
  \includegraphics[width=0.35\textwidth]{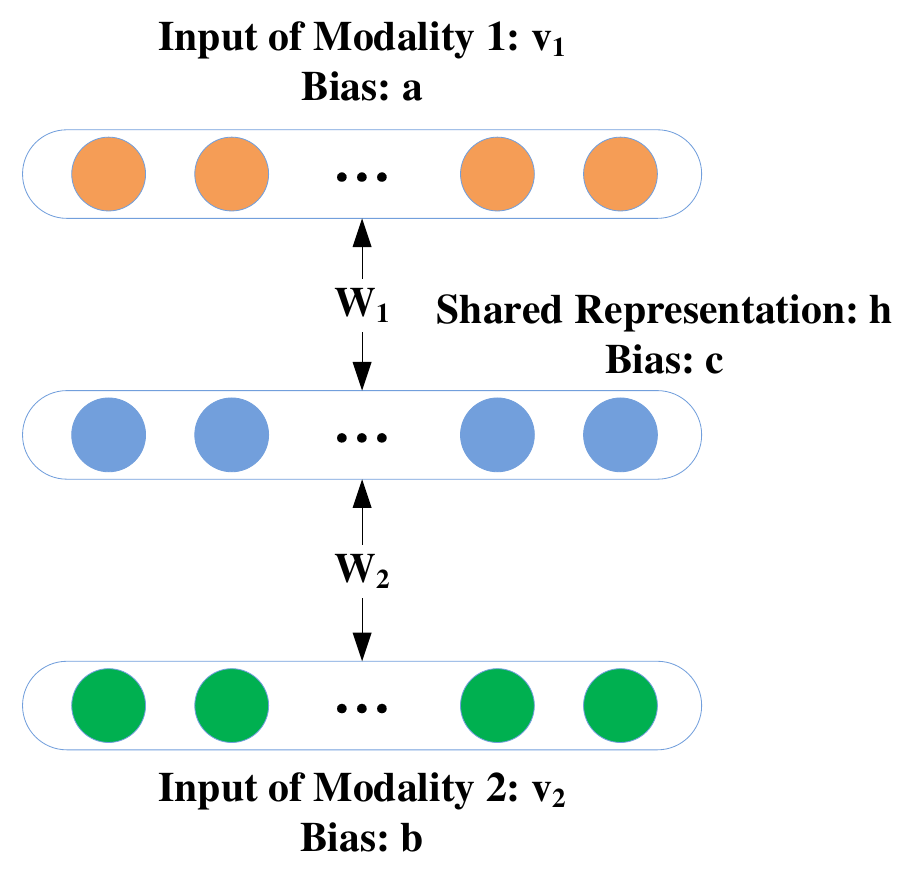}\\
  \caption{A multi-modal RBM that modeling the joint distribution of face images in two modalities. The hidden layer in the model can be seen as a shared representation of the two input modalities.}
  \label{fig:multi-modal-RBM}
\end{figure}

\cite{Srivastava-NIPS-2012} constructed a multi-modal RBM to model the relationship between image and text by combining a Gaussian RBM and Replicated Softmax RBM. For heterogenous face recognition problem, we use two Gaussian RBM to model the relationship between face data in two modalities. The structure of our model is shown in Figure~\ref{fig:multi-modal-RBM}. Its energy function is given by:
\begin{equation}
\begin{split}
E(\hat{\bfv}_1, \hat{\bfv}_2, \bfh; \boldsymbol\theta) =
& \frac{1}{2} (\hat{\bfv}_1 - \bfa)^T (\hat{\bfv_1} - \bfa) + \\
& \frac{1}{2} (\hat{\bfv}_2 - \bfb)^T (\hat{\bfv_2} - \bfb) - \\
& \bfc^T \bfh  - \hat{\bfv}_1^T \bfW_1 \bfh - \hat{\bfv}_2^T \bfW_2 \bfh,
\end{split}
\end{equation}
where $\hat{\bfv}_1$ and $\hat{\bfv}_2$ are face images in two modalities; $\bfW_1$ and $\bfW_2$ are weights matrix for each modality respectively. The joint distribution over $\hat{\bfv}_1$, $\hat{\bfv}_2$, and $\bfh$ can be calculated based on the energy function, as similar as Equ.~(\ref{equ:energy-to-prob}).

Given the normalized training data $\hat{\bfv}_1$ and $\hat{\bfv}_2$, we can learn the parameters $\boldsymbol\theta$. Then, the trained multi-modal RBM can be used flexibly, such as
\begin{enumerate}
\item generating missing modality by sampling from conditional distribution $P(\hat{\bfv}_1 | \hat{\bfv}_2)$,
\item fusing two modalities by sampling from $P(\bfh | \hat{\bfv}_1, \hat{\bfv}_2)$,
\item inferring shared representation by sampling from $P(\bfh | \hat{\bfv}_1)$ and $P(\bfh | \hat{\bfv}_2)$ respectively.
\end{enumerate}
Due to the experience in heterogeneous literature~\cite{Lin-ECCV-06}\cite{Lei-TIFS-2012}\cite{Zhang-CVPR-2011}, this paper uses it for shared representation inference, which transforms the heterogenous data into a common space. For the details of multi-modal RBM learning and inference, please refer to \cite{Salakhutdinov-JMLR-2009}\cite{Srivastava-NIPS-2012}.

\section{Learning Shared Representation}
\label{sec:sr}

\subsection{Framework}

\begin{figure}
  \centering
  \includegraphics[width=0.85\textwidth]{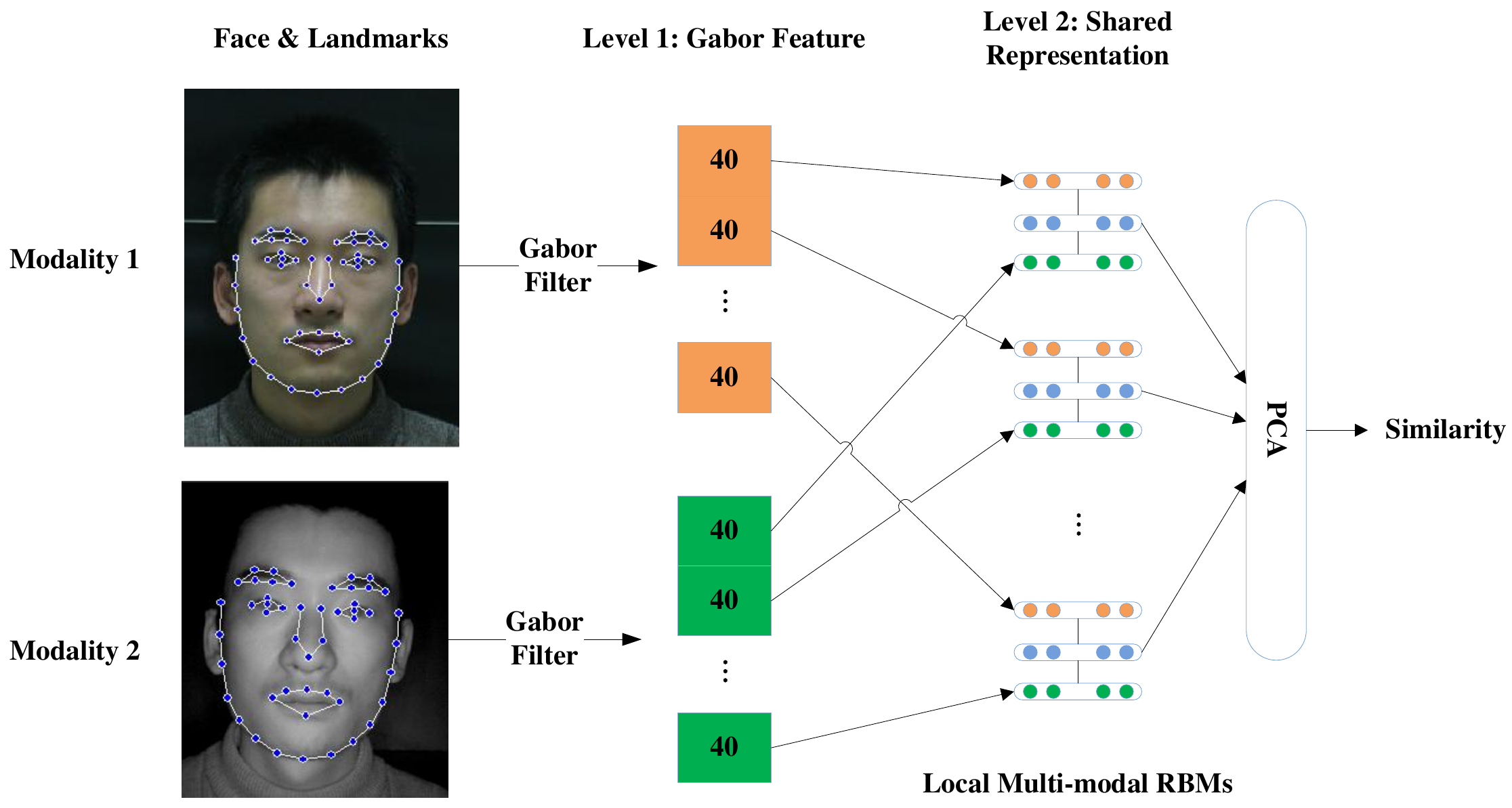}\\
  \caption{The proposed framework for heterogeneous face recognition by combining traditional face recognition modules and local RBMs.}
  \label{fig:framework}
\end{figure}

The core of heterogeneous face recognition is modeling the relationship between different modalities and meanwhile reserving the discriminative information. To this end, we propose a framework for heterogenous face recognition by incorporating RBM into the traditional face recognition pipeline. The flowchart of the framework is shown in Figure~\ref{fig:framework}, in which the heterogenous face images are illustrated by NIR and VIS for example. First, Gabor features are extracted at many facial points for two modalities respectively. Based on the Gabor features, a series of local RBMs are used to learn the shared representation of two modalities for each facial point. All local shared representations are then concatenated and processed by PCA. Finally the similarity of these modality-free features can be evaluated by Cosine metric.

The proposed framework has following advantages:
\begin{enumerate}
 \item Local Gabor feature is the mainstream in face recognition, which has strong discriminative ability and is robust to variations;
 \item We learn the shared representation locally because the modality gap is smaller in local region, and low dimensional data is more efficient for computation and easier to prevent overfitting;
 \item PCA can remove the redundance and heterogeneity further in holistic scale.
\end{enumerate}

The details of each step in Figure~\ref{fig:framework} will be discussed in the following subsections.

\subsection{Level 1 Representations}

The task of level 1 is to extract discriminant and robust features for each modality. Recently, local features based on facial points achieved excellent performance in face recognition~\cite{Chen-CVPR-2013}\cite{Yi-CVPR-2013}, especially in unconstrained face recognition, \eg LFW~\cite{LFWTech}. Although the face images in heterogeneous databases are both near frontal, facial points are still be used to deal with the small pose variations.

\begin{figure}
  \centering
  \includegraphics[width=0.5\textwidth]{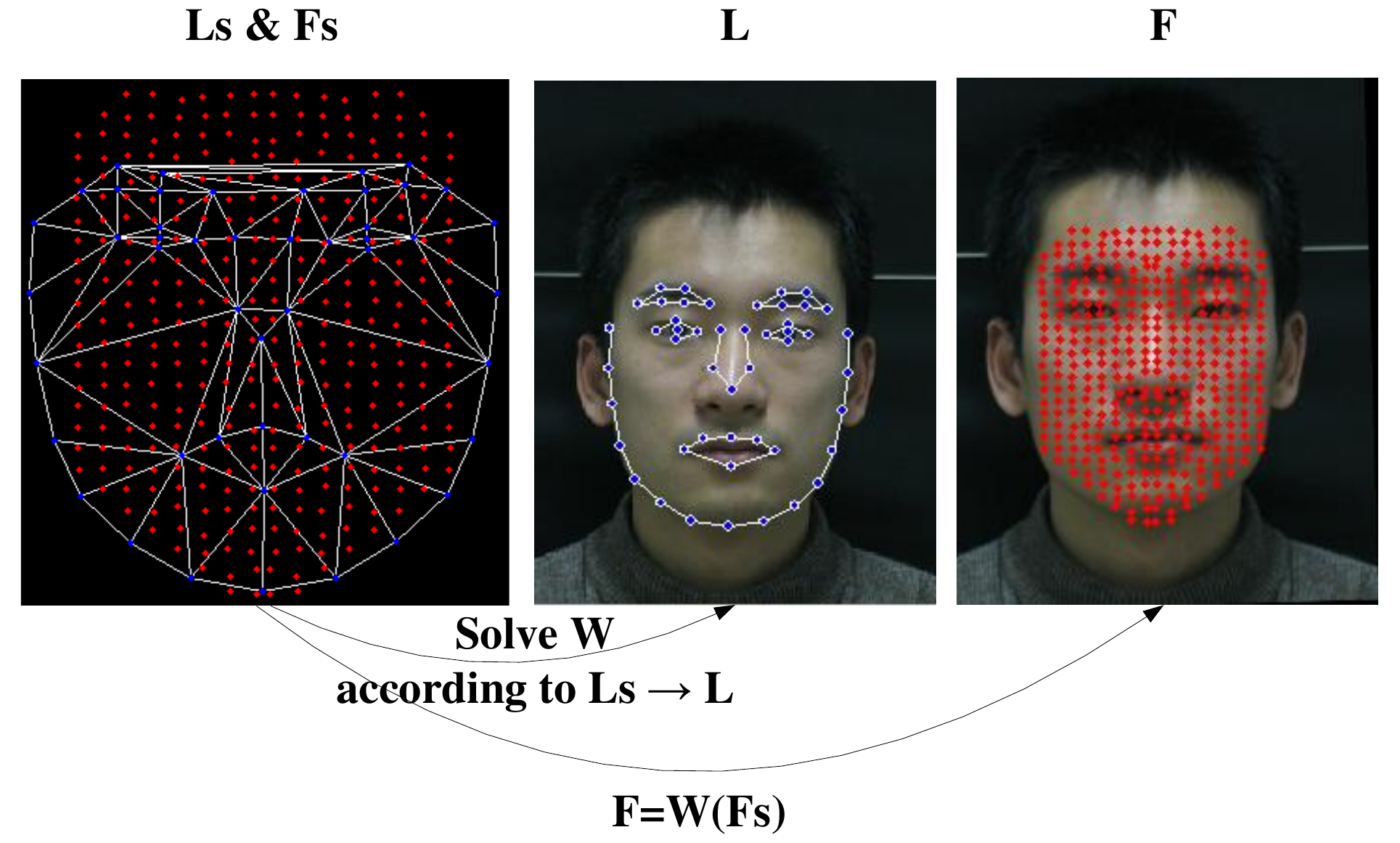}\\
  \caption{The warping process of facial points. Left: Standard landmarks $\bfL_s$ (blue dots) and facial points $\bfF_s$ (red dots). Middle: A face image and its corresponding landmarks $\bfL$. Right: the warped facial points $\bfF$ for the input image.}
  \label{fig:warping}
\end{figure}

As shown in Figure~\ref{fig:warping}, a standard set of facial points $\bfF_s$ are defined for feature extraction and another 48 landmarks $\bfL_s$ are defined for alignment, similar to \cite{Yi-CVPR-2013}. Given a face image, we need put the facial points to the right place on it. \cite{Yi-CVPR-2013} used a fast 3DMM model to do this work. For simplicity, this paper uses RBF warping~\cite{Arad-CGF-1995} to transform the standard facial points to the face image. The warping process is shown in Figure~\ref{fig:warping}. Given the landmarks $\bfL$ of the input image, a warping function $W$ can be solved based on $\bfL_s$ and $\bfL$. Then the warped facial points are calculated by $\bfF = W(\bfF_s)$. We can see that the facial points can fit the input image well. The deformation factor of RBF warping is set to $0.1 \times$ ``eye distance''.

At the warped $176\times2$ facial points $\bfF$, local features are extracted by a Gabor wavelet described in \cite{Okada-Bochum-98}. The space of Gabor wavelet is sampled in 8 orientations and 5 resolutions, thus giving $5\times8 = 40$ features for each facial point. Since the facial points are defined in a symmetric way, the features are grouped in left and right halves. Thus we get two feature vectors with $40\times176$ dimensions for each face image. Note that the facial symmetry trick has been used in many papers~\cite{Thomas-BMVC-2012}\cite{Li-CVPRW-2009}, which can augment the dataset and improve the computation efficiency.

\subsection{Level 2 Representations}

The task of level 2 is to build the relationship between two modalities. Previous work~\cite{Yang-FG-2008} has proven that the local relationship is easier to learn than holistic relationship, therefore we use local RBM to learn shared representation for each facial point. The structure of the RBMs is 40-80-40, including two input linear layers and a logistic hidden layer. Because the dimension of input of the RBM is very low, no sparse penalty and weight decay are used.

Existing methods, such as CSR~\cite{Lei-TIFS-2012}, CITE~\cite{Zhang-CVPR-2011} and their nonlinear versions, often learn the relationship in supervised and discriminative way. Different from them, RBM learns the joint distribution of the two modalities in a generative way, so RBM is less affected by overfitting. As described in \cite{Huang-CVPR-2012}, the relationships between modalities are not stationary with respect to the location in image, so we use many local RBMs to model them respectively, instead of one holistic RBM.

The level 1 features of two modalities are sent to 176 local RBMs, and their parameters are learned by using mean-field inference and an MCMC procedure described in \cite{Srivastava-NIPS-2012}. In the training stage, the batch size is set to 10 and the number of batches is set to 50000. After the training is completed, we can infer the shared representations of two modalities by sampling from $P(\bfh | \hat{\bfv}_1)$ and $P(\bfh | \hat{\bfv}_2)$. While sampling from $P(\bfh | \hat{\bfv}_1)$, we treat $\hat{\bfv}_2$ and $\bfh$ as missing data and initialize them randomly, then generate the hidden representation $\bfh$ by alternating Gibbs sampler~\cite{Srivastava-NIPS-2012}. The hidden representation of another modality can be generated in a similar way. The activation probabilities of the hidden layer are called the shared representation of heterogenous face images. The size of shared representation of a half face is $80\times176$.

\subsection{Cross-modal Matching}

After the heterogeneity has been removed in local regions, the heterogeneity over holistic face still exists. As described in \cite{Li-CVPRW-2013}, PCA can capture the heterogeneity in its first several principle components, so we use PCA to process the feature in a holistic way. First, the 176 local representations are concatenated into a vector (the dimension is $80\times176=14080$) and the first several principle components of PCA are then removed. The number of removed components is tuned on the training set or development set. To this stage, the features of two modalities are actually transformed into a common space. Their similarity is calculated by Cosine metric. The similarity of two face halves are fused by sum rule.

We have also tried to learn a discriminative distance metric by LDA and Metric Learning based on the shared representations, but got worse results than PCA. The reason may be due to the limited data. We believe that the supervised methods will outperform PCA after having larger database in the future.

\section{Experiments}
\label{sec:exp}

To illustrate the superior performance of the proposed method, we take Sketch-Photo and NIR-VIS face recognition problems to conduct experiments. The results on three popular databases, CUFS~\cite{Wang-TPAMI-2009}, CASIA HFB~\cite{Li-CVPRW-2009} and CASIA NIR-VIS 2.0~\cite{Li-CVPRW-2013}, all outperform the current state-of-the-arts significantly.

\subsection{Sketch to Photo}

For Sketch-Photo problem, the CUFS database is used. The photos in CUFS come from three sources: 188 faces from CUHK student dataset, 123 faces from AR, and 295 faces from XM2VTS. Their corresponding sketches are drawn by an artist. In total, CUFS contains 606 subjects, 1 photo and 1 sketch per subject. As suggested in \cite{Zhang-CVPR-2011}, the database is split into 306 training subjects and 300 testing subjects. To get unbias results, the process repeats 10 times, and generates 10 splits. The mean and standard deviation of recognition rate of the 10 splits are reported. In the testing phase, photo is used as gallery and sketch is used as probe.

Because many good results has been reported on CUFS, it is considered as a relatively easy database~\cite{Klare-TPAMI-2011}. For this simple experiment, the proposed method can work well without RBM, just using Gabor feature and PCA. Every photos and sketches are processed by facial points detection, Gabor feature extraction, PCA and Cosine matching. Only using Gabor feature, we can achieve comparable result to state-of-the-art methods, \ie Rank1=99.50\%. By removing the first 20 principle components, the differences between photo and viewed sketch are removed completely, and enough identify information are reserved for classification. Without RBM, we get 100\% recognition rate and outperform other compared methods. The comparisons are shown in Table~\ref{tbl:cufs-results}.

\begin{table}
  \centering
  \caption{Rank1 recognition rates and VR@FAR=0.1\% of various methods on CUFS.}
  \label{tbl:cufs-results}
  \begin{tabular}{|c|c|c|}
    \hline
     & Rank1 & VR \\
    \hline
    Gabor & $99.50\pm0.39\%$ & $94.70\pm1.2\%$ \\
    \hline
    Gabor + Remove 20 PCs & $100\pm0\%$ & $100\pm0\%$ \\
    \hline
    MRF+RS-LDA~\cite{Wang-TPAMI-2009} & $96.30\%$ & N/A \\
    \hline
    LFDA~\cite{Klare-TPAMI-2011} & $99.47\%$ & N/A \\
    \hline
    CITE~\cite{Zhang-CVPR-2011} & $99.87\%$ & N/A \\
    \hline
  \end{tabular}
\end{table}

\subsection{NIR to VIS}

To illustrate the performance of our method further, two more difficult experiments are conducted on the CASIA HFB and NIR-VIS 2.0 databases.

CASIA HFB contains 2095 VIS and 3002 NIR face images from 202 subjects. We follow the evaluation protocol in \cite{Klare-ICPR-2010} that selects 102 subjects for training and the other 100 subjects for testing. Similar to the previous experiment, the random selection is repeated 11 times. The first split (View 1) is used to tune the parameters of algorithm, and the other 10 splits (View 2) are used to report the performance.

CASIA NIR-VIS 2.0 is an upgraded version of HFB, the images in which are captured using the same devices as HFB, but has larger scale, contains more variations in pose, facial expression and age. CASIA NIR-VIS 2.0 is more close to practical applications. This database has standard evaluation protocols, so we use them directly.

In these two experiments, VIS face images are used as gallery and NIR face images are used as probe.

\subsubsection{CASIA HFB}

First, the same framework in the previous experiment is applied on HFB, but just get moderate result. The reason may be that the first 20 principle components cannot capture the full difference between modalities. To verify this, we tune the number of removed principle components on View 1 carefully, but the VR (Verification Rate) cannot increase anymore. Thus we think the heterogeneity and discriminative information are coupled tightly and need to be dealt with in low level by RBM.

\begin{figure}
  \centering
  \includegraphics[width=0.5\textwidth]{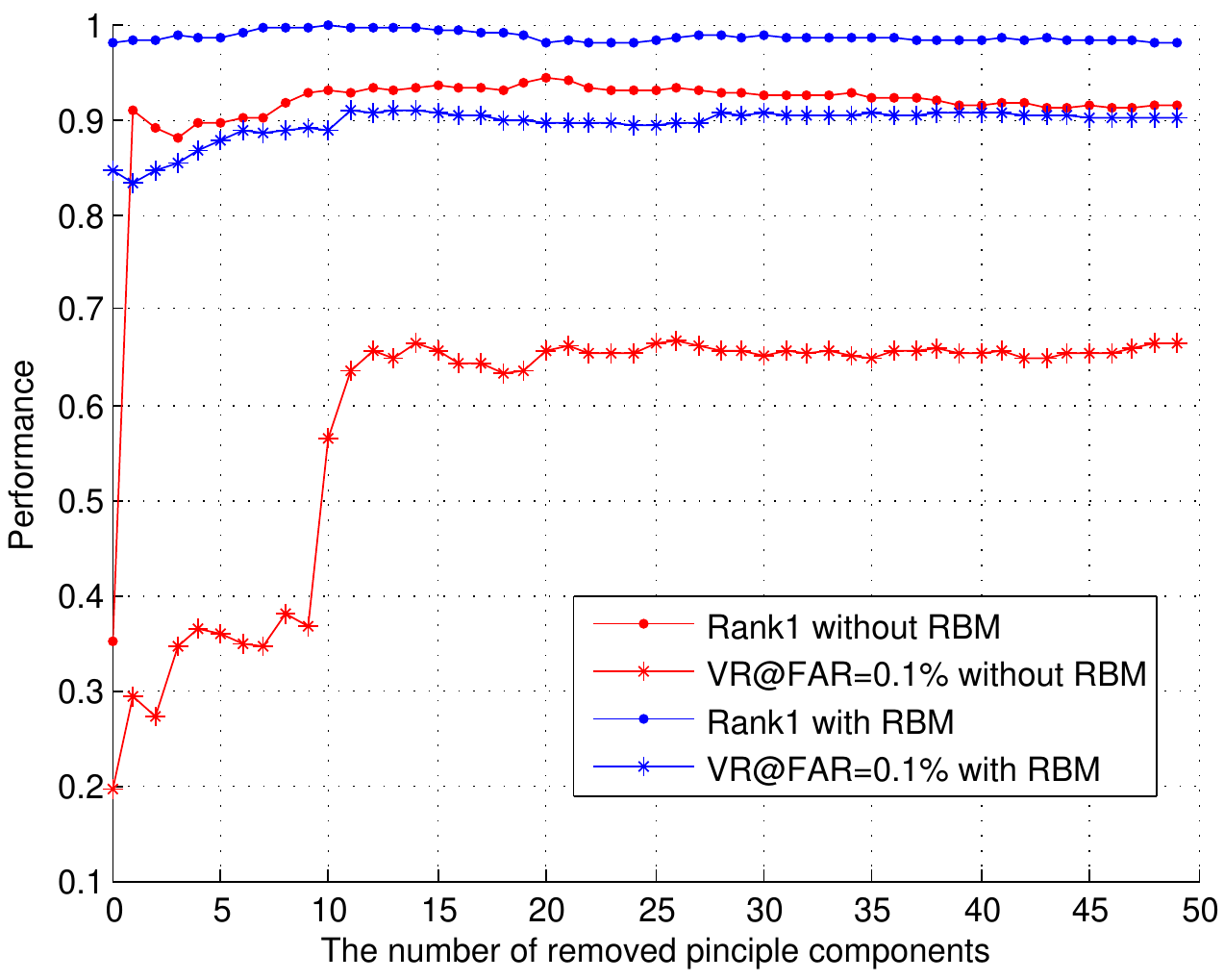}\\
  \caption{The relationship between Rank1, VR and the number of removed principle components on View 1 of CASIA HFB. And the comparison curves of our method with/without RBMs.}
  \label{fig:hfb-npc}
\end{figure}

After introducing the RBM, the performance of our method increases significantly. As shown in Table~\ref{tbl:hfb-results}, the VR@FAR=0.1\% is improved from 71.70\% to 92.25\% and the deviation is also reduced remarkably. Meanwhile, the optimal number of removed principle components drops from 20 to 11 (see Figure~\ref{fig:hfb-npc}), which indicates that the modality-free representations are successfully learned by local RBMs.

Compared to other methods in \cite{Klare-ICPR-2010} and \cite{Lei-TPAMI-2013}, the Rank1 and VR of our method are obviously higher. The SR (Sparse Representation) in \cite{Klare-ICPR-2010} used the whole gallery to optimize the matching process, which has been proved can improve performance, especially in terms of ROC curve. For example, the VR of our method can be improved from 92.25\% to 96.33\% by using z-score normalization~\cite{Jain-PR-05}. Because in face verification applications we cannot obtain the whole gallery, we just report the results without using the whole gallery. By fusing two classifiers and a commercial face recognition SDK, the VR of NN+SR+Cognitec~\cite{Klare-ICPR-2010} outperforms ours slightly. The reported performance of P-RS~\cite{Klare-TPAMI-2013} is better than ours, but it is trained on larger training set. And P-RS is surely slower than our method because it's based on kernel similarities.

\begin{table}
  \centering
  \caption{Rank1 recognition rates and VR@FAR=0.1\% of various methods on View 2 of CASIA HFB.}
  \label{tbl:hfb-results}
  \begin{tabular}{|c|c|c|}
    \hline
     & Rank1 & VR \\
    \hline
    Gabor & 59.47$\pm6.72\%$ & $33.51\pm5.70\%$ \\
    \hline
    Gabor + Remove 20 PCs& $94.87\pm1.72\%$ & $71.70\pm6.42\%$ \\
    \hline
    Gabor + RBM + Remove 11 PCs & $99.38\pm0.32\%$ & $92.25\pm1.68\%$ \\
    \hline
    NN~\cite{Klare-ICPR-2010} & $88.8\%$ & $48.78\pm3.87\%$ \\
    \hline
    SR~\cite{Klare-ICPR-2010} & $93.4\%$ & $77.56\pm2.96\%$ \\
    \hline
    NN+SR~\cite{Klare-ICPR-2010} & $92.2\%$ & $79.05\pm4.48\%$ \\
    \hline
    Cognitec~\cite{Klare-ICPR-2010} & $93.8\%$ & $85.62\pm2.17\%$ \\
    \hline
    NN+SR+Cognitec~\cite{Klare-ICPR-2010} & $97.6\%$ & $93.45\pm0.96\%$ \\
    \hline
    C-DFD~\cite{Lei-TPAMI-2013} & $92.2\%$ & $65.5\%$ \\
    \hline
    P-RS~\cite{Klare-TPAMI-2013}\tablefootnote{133 subjects for training, 67 subjects for testing} & - & $95.8\pm6.15\%$ \\
    \hline
  \end{tabular}
\end{table}

\paragraph{Global, Convolutional and Local RBMs}

The layer in neural network has three popular styles: fully collected layer, locally collected layer with shared weights (convolutional) and locally collected layer with unshared weights (local). For RBM, we call them as global, convolutional and local RBMs. To illustrate the advantages of local RBMs, we plug them into our framework and compare their performances on View 1 of HFB, the information of which are shown in Table~\ref{tbl:RBMs}. The architecture of convolutional and local RBMs are all 40-80-40. Limited by the memory of our Geforce GTX670 GPU, the hidden layer of global RBM only uses 3520 units.

The complexity of the three kinds of RBMs are global $>$ local $>$ convolutional. Generally, complex models are easier to overfit to the training set and simple models are prone to underfitting. The results in Table~\ref{tbl:RBMs} prove this point very well. The global RBM just performs well on the training set and the convolutional RBM performs moderately both on training and testing set. Among these models, the local RBMs obtain the best trade-off between complexity and generalization. Maybe the locality of connection and weight sharing can be fine-tuned further to get better results, but we leave this work to the future.

\begin{table}
  \centering
  \caption{The comparison of Global, Convolutional and Local RBMs on View 1 of CASIA HFB. The 3rd column is VR@FAR=0.1\% on the training set of View 1. The 4th column is VR@FAR=0.1\% on the testing set of View 1.}
  \label{tbl:RBMs}
  \begin{tabular}{|c|c|c|c|}
    \hline
     & Architecture  & VR (Train) & VR (Test)  \\
    \hline
    Global & 7040-3520-7040 & 99.94\% & 1.549\% \\
    \hline
    Conv. & 40-80-40 & 73.31\% & 71.79\% \\
    \hline
    Local & $176\times$(40-80-40) & 99.45\% & 90.85\% \\
    \hline
  \end{tabular}
\end{table}

\paragraph{Parameter Tuning}

As discussed above, the number of removed principle components greatly affect the performance of our method. Generally, if the difference between modalities is bigger, we need drop more principle components. However, there are also some identity information existing in these components, so we should find a trade-off. Figure~\ref{fig:hfb-npc} shows the relationship between the performance and the number of removed principle components on View 1. From the figure we can see that the performance of our method without RBMs are affected drastically by the first several principle components. But after using RBMs, the curves become smoother and quick to reach the optimal point. Therefore, we set the number of removed PCs to 20 when without RBMs and set the number to 11 when with RBMs.

\paragraph{Failure Cases}

\begin{figure}
  \centering
  \includegraphics[width=0.5\textwidth]{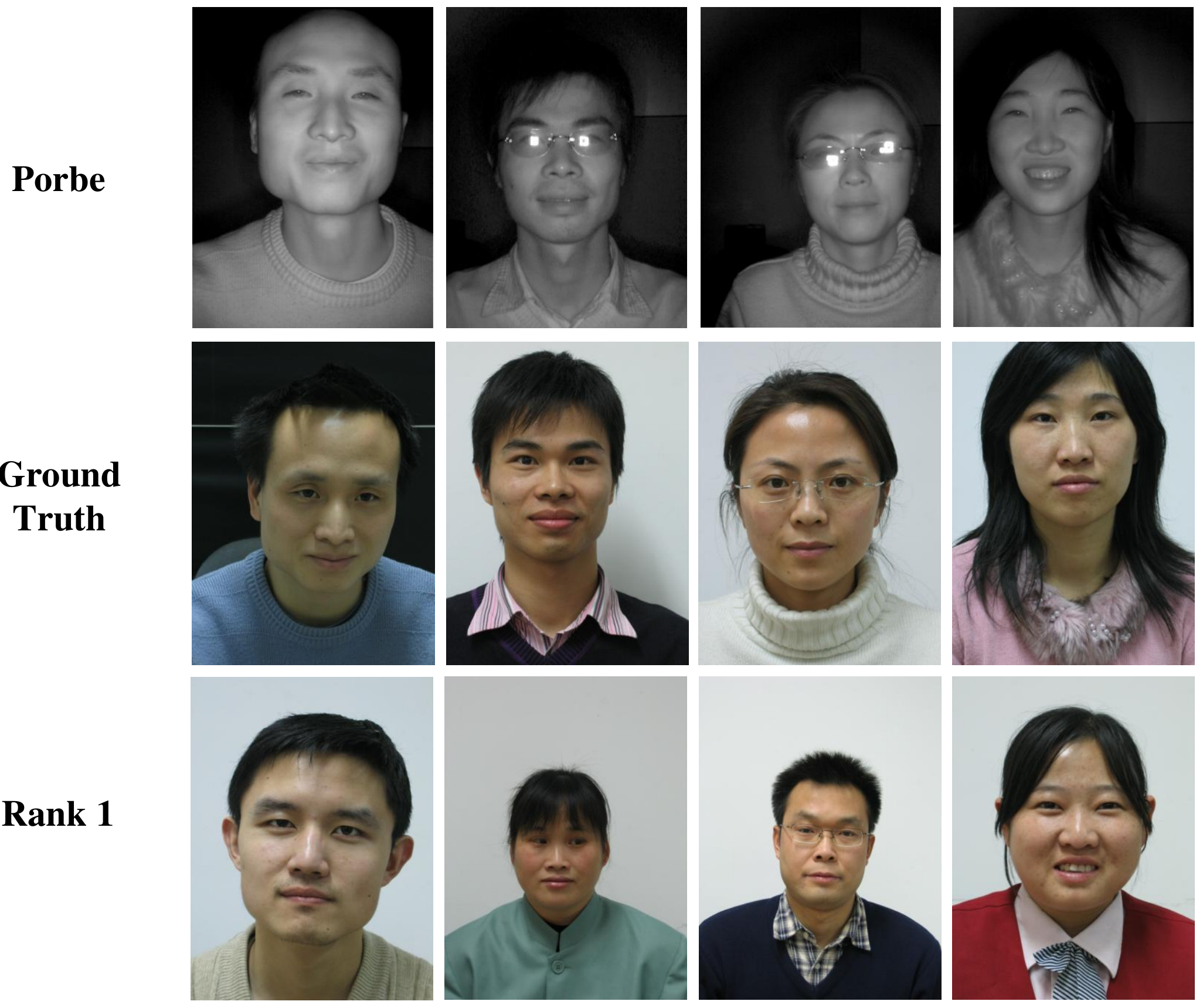}\\
  \caption{The four failure cases on View 1 of CASIA HFB database. The first row are the NIR probe face images. The second row are the corresponding VIS face images of the first row. The third row are the retrieved Rank1 results of our method.}
  \label{fig:failure-cases}
\end{figure}

Although the Rank1 recognition rate of our method is very high, there are still four failure cases on View 1 of HFB, which are shown in Figure~\ref{fig:failure-cases}. From the figure we can see that the four NIR probe images both have obvious variations in pose, specular reflectance on eyeglasses and expression. Even in traditional paradigm these factors heavily degrade the performance of face recognition, and they are more difficult to solve when coupling with spectrum variations.

\subsubsection{CASIA NIR-VIS 2.0}

CASIA NIR-VIS 2.0 is a more challenging and practical database than the above two databases. The process of this experiment is as same as HFB, by first tuning the parameters on View 1 and then reporting results on View 2. From the results (Table~\ref{tbl:nir-vis-2-results}) we can see that the Rank1 and VR on NIR-VIS 2.0 drop 10-20\% compared to HFB. On this database, the improvements bringed by removing the first PCs and RBM are still obvious, about 40\% and 10\% respectively.
Because NIR-VIS 2.0 is a new database, we just list the baseline in \cite{Li-CVPRW-2013} for comparison.

\begin{table}
  \centering
  \caption{Rank1 recognition rates and VR@FAR=0.1\% of various methods on View 2 of CASIA NIR-VIS 2.0.}
  \label{tbl:nir-vis-2-results}
  \begin{tabular}{|c|c|c|}
    \hline
     & Rank1 & VR \\
    \hline
    Gabor & $36.18\pm2.56\%$ & $33.37\pm2.29\%$ \\
    \hline
    Gabor + Remove 20 PCs & $75.54\pm0.75\%$ & $71.40\pm1.21\%$ \\
    \hline
    Gabor + RBM + Remove 11 PCs & $86.16\pm0.98\%$ & $81.29\pm1.82\%$  \\
    \hline
    PCA+Sym+HCA~\cite{Li-CVPRW-2013} & $23.7\pm1.89\%$ & $19.27\%$ \\
    \hline
  \end{tabular}
\end{table}

\section{Conclusion}
\label{sec:con}

This paper proposed a novel framework for heterogeneous face recognition by combing RBM and the popular modules from face recognition. Because of its unsupervised nature, the framework is not prone to overfitting problem, and work well on many challenging heterogeneous face databases. Based on Gabor features, the modality-free shared representations were first learned successfully in low level by many local RBMs, and further processed by PCA in high level. The proposed framework performed perfectly on the CUFS database and outperformed state-of-the-art methods significantly on the CASIA HFB and NIR-VIS 2.0 databases. Moreover, all experimental results illustrated the success of local RBMs to learn the shared representations. The future work will be conducted in two directions: (1) by stacking many multi-modal RBMs to learn high level representations; (2) exploring the way to fine tune the model with identity information.



\bibliographystyle{splncs}
\bibliography{vision}

\end{document}